\pdfoutput=1

\documentclass[11pt]{article}

\usepackage{acl}

\usepackage{times}
\usepackage{latexsym}

\usepackage{float}
\usepackage{graphicx}
\usepackage{xspace}
\usepackage{subcaption}
\usepackage{makecell}
\usepackage{booktabs}
\usepackage{multirow}
\usepackage{xcolor}
\usepackage{amsmath, amsfonts}
\usepackage{amssymb}
\usepackage{bm}
\usepackage{pdflscape}
\usepackage{afterpage}
\usepackage{enumitem}
\usepackage{pifont}

\usepackage{dirtytalk}

\newcommand{\RN}[1]{%
  \textup{\uppercase\expandafter{\romannumeral#1}}%
}
\usepackage{threeparttable}

\usepackage{dsfont}

\usepackage[T1]{fontenc}

\usepackage[utf8]{inputenc}

\usepackage[english]{babel}
\usepackage{amsthm}
\newtheorem{innercustomgeneric}{\customgenericname}
\providecommand{\customgenericname}{}
\newcommand{\newcustomtheorem}[2]{%
  \newenvironment{#1}[1]
  {%
   \renewcommand\customgenericname{#2}%
   \renewcommand\theinnercustomgeneric{##1}%
   \innercustomgeneric
  }
  {\endinnercustomgeneric}
}

\newcustomtheorem{customthm}{Theorem}
\newcustomtheorem{customlemma}{Lemma}

\DeclareMathOperator{\argmin}{arg\,min}

\usepackage{nccmath}

\usepackage{microtype}

\newcommand{\thicktilde}[1]{\mathbf{\tilde{\text{$#1$}}}}
\allowdisplaybreaks
\title{Distantly Supervised Named Entity Recognition via Confidence-Based Multi-Class Positive and Unlabeled Learning}

\author{Kang Zhou,\hspace{2mm} Yuepei Li,\hspace{2mm} Qi Li \\
  Department of Computer Science, Iowa State University \\
  \texttt{\{kangzhou, liyp0095, qli@iastate.edu\}} \\}

\begin{document}
\maketitle
\begin{abstract}
In this paper, we study the named entity recognition (NER) problem under distant supervision. Due to the incompleteness of the external dictionaries and/or knowledge bases, such distantly annotated training data usually suffer from a high false negative rate. To this end, we formulate the Distantly Supervised NER (DS-NER) problem via Multi-class Positive and Unlabeled (MPU) learning and propose a theoretically and practically novel CONFidence-based MPU (\textbf{Conf-MPU}) approach. To handle the incomplete annotations, Conf-MPU consists of two steps. First, a confidence score is estimated for each token of being an entity token. Then, the proposed Conf-MPU risk estimation is applied to train a multi-class classifier for the NER task. Thorough experiments on two benchmark datasets labeled by various external knowledge demonstrate the superiority of the proposed Conf-MPU over existing DS-NER methods.
\end{abstract}

\begingroup\makeatletter\def\f@size{10}\check@mathfonts
\def\maketag@@@#1{\hbox{\m@th\large\normalfont#1}}%

\section{Introduction} \label{introduction}
Named Entity Recognition (NER) aims to detect entity mentions from text and classify them into predefined types. It is a fundamental task in information extraction and many other downstream tasks \cite{gabor2018semeval, luan2017scientific, giorgi2019end}.
However, the necessity of extensive human efforts to annotate a large amount of training data imposes restrictions on the state-of-the-art supervised deep learning methods, especially in professional fields.

To address this problem, distantly supervised methods spring up, which aim to train NER models using automatically annotated training data based on external knowledge such as dictionaries and knowledge bases. Observed by previous DS-NER methods \cite{shang2018learning, peng2019distantly}, distant labels provided by reliable dictionaries are usually of high precision. However, such distant labeling suffers a major drawback --- incomplete labeling. This is due to the fact that most existing dictionaries and knowledge bases have limited coverage on entities. Hence simply treating unlabeled samples (\textit{i.e.}, unmatched tokens) as negative ones will introduce a high false negative rate (\textit{e.g.}, \say{neutropenia} in Figure \ref{fig:figure_1}) compared with human-annotated training data, and further mislead a supervised NER model to overfit to false negative samples and seriously affect its recall.
\begin{figure}[t]
    \centering
    \includegraphics[width=0.48\textwidth]{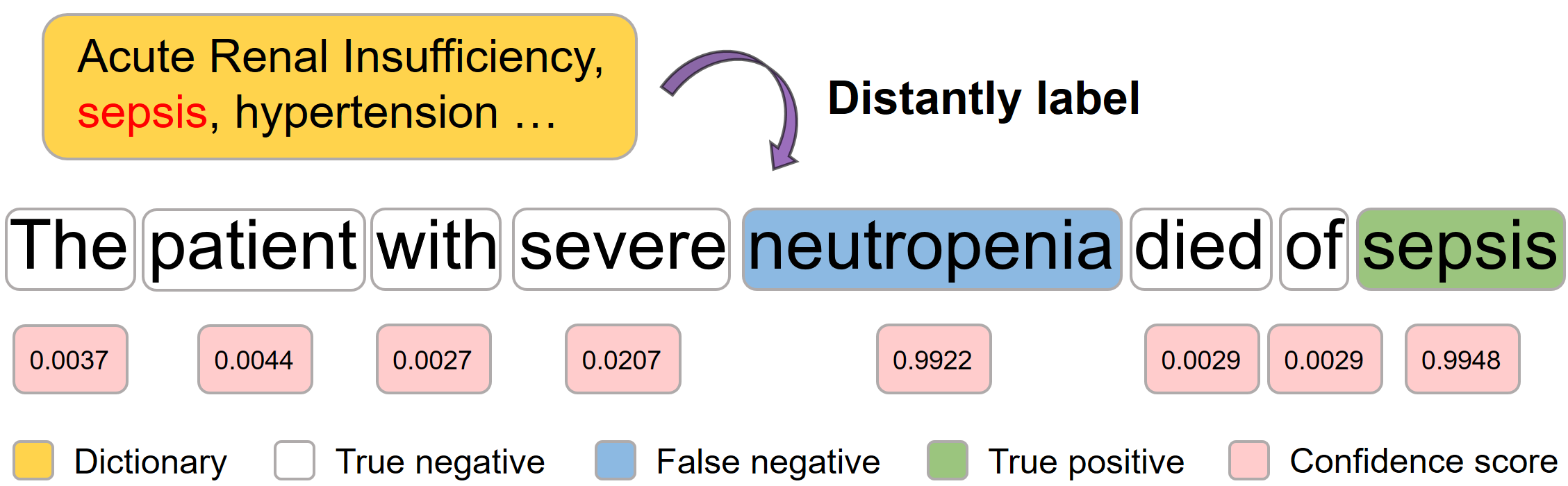}
    \caption{Distant labeling example for the entity type of \texttt{Disease} using a dictionary.}
    \label{fig:figure_1}
    \vspace{-5mm}
\end{figure}

To tackle this challenge, several DS-NER methods have been put forward recently for the incompletely annotated data. One line of work focuses on designing deep learning architectures that can cope with the training data with high false negative rates to partially alleviate the impact of the defective distant annotations \cite{shang2018learning, liang2020bond}. Another line of work applies partial conditional random field (CRF) to assign unlabeled samples with all possible labels and maximize the overall probability \cite{mayhew2019named, cao2019low, yang2018distantly, shang2018learning}.

Recently, binary Positive and Unlabeled (PU) learning is applied to DS-NER tasks \cite{peng2019distantly}. PU learning performs classification using only limited labeled positive data and unlabeled data, thus naturally suitable for handling distant supervision, where external knowledge often has a limited coverage on positive samples.
However, binary PU learning has several drawbacks in real DS-NER tasks. It applies the one-vs-all strategy to convert a multi-class classification problem into multiple binary classification problems, and thus suffers from two weaknesses. First, it is not efficient, especially in the case where there are many entity types. For a NER task with $n$ entity types, $n$ binary classifiers need to be trained. Second, the scale of predicted confidence values may differ among those binary classifiers, which may not guarantee a mutually beneficial inference for the final prediction \cite{bishop2006pattern}.

Furthermore, the PU learning theory is built on a fundamental assumption of data distribution that unlabeled data can accurately reveal the overall distribution (\textit{i.e.}, the marginal distribution of the target field) \cite{bekker2020learning}. In DS-NER tasks, the distantly annotated training data may not fit the assumption well: It depends on the coverage of used dictionaries or knowledge bases on the entities. Our empirical studies validate that violation of this assumption can significantly impact the performance of PU learning. 


To address these challenges in DS-NER tasks and PU learning, we propose a CONFidence-based Multi-class Positive and Unlabeled (\textbf{Conf-MPU}) learning framework.
The proposed Conf-MPU can handle different levels of false negative rates brought by dictionaries of various coverage and does not overfit to the distantly labeled training data. It consists of two steps. Specifically, given the distantly labeled training data, we first carry out a token-level binary classification to estimate the confidence score (a probability value in $[0, 1]$) of a token being an \textit{entity token} (\textit{i.e.}, a token of a named entity). Then, we perform the NER classification using a neural network model with the proposed Conf-MPU risk estimator, which incorporates the confidence scores obtained from the first step in the risk estimation, to alleviate the impact of annotation imperfection. It is worth noting that the two-step strategy of Conf-MPU needs to train only two classifiers for any DS-NER tasks with arbitrary number of entity types, which is more efficient than previous binary PU learning. 
 
In summary, our main contributions are:
\begin{itemize}[wide, nosep]
\item We propose Conf-MPU, a theoretically and practically novel approach for the DS-NER task. Conf-MPU enriches the PU learning theory with solid theoretical analysis.
\item We verify that the practical use of traditional PU learning is subject to its theoretical assumption, which can be released by Conf-MPU. As far as we know, this is the first work specially dealing with such a practical problem.
\item We empirically demonstrate that Conf-MPU with a two-step strategy can significantly alleviate the impact of incomplete annotations during the model training and outperform the state-of-the-art DS-NER methods on benchmark datasets.
\end{itemize}

\section{Preliminaries}
In this section, we briefly review the risk formulations of standard supervised learning and PU learning in the binary classification setting. 

\subsection{Standard Binary Supervised Learning}
Suppose that the data follow an unknown probability distribution with density $p(\boldsymbol{x}, y)$. Let $\boldsymbol{x} \in \mathcal{X} \subseteq \mathbb{R}^d$ and $y \in \mathcal{Y} = \{0, 1\}$, where $0$ and $1$ indicate negative and positive classes, respectively. The goal is to learn a decision function $f: \mathcal{X} \rightarrow \mathcal{Y}$ by minimizing the expected classification risk:
\begin{align} \label{eq1}
        \mathrm{R}(f) = {} &\pi \mathrm{R}_{\rm P}^{+}(f) + (1 - \pi) \mathrm{R}_{\rm N}^{-}(f). 
\end{align}
In this function, $\pi = p(y=1)$ is the prior of the positive class. $\mathrm{R}_{\rm P}^{+}(f) = \mathbb{E}_{\boldsymbol{x}\sim p(\boldsymbol{x}\mid y=1)} \left[\ell(f(\boldsymbol{x}), 1)\right]$ and $\mathrm{R}_{\rm N}^{-}(f) = \mathbb{E}_{\boldsymbol{x}\sim p(\boldsymbol{x}\mid y=0)} \left[\ell(f(\boldsymbol{x}), 0)\right]$ denote the expected classification risks on the positive and negative classes, respectively, where $\mathbb{E}$ denotes expectation and its subscript indicates the data distribution on which the expectation is computed, and the loss function is represented by $\ell$.

In supervised learning setting, we are given both labeled positive and negative data that are sampled independently from $p_{\rm{P}}(\boldsymbol{x}) = p(\boldsymbol{x} \mid y = 1)$ and $p_{\rm{N}}(\boldsymbol{x}) = p(\boldsymbol{x} \mid y = 0)$ as $\mathcal{X}_{\rm{P}} = \{\boldsymbol{x}^{\rm{P}}_{j}\}^{n_{\rm{P}}}_{j=1}$ and $\mathcal{X}_{\rm{N}} = \{\boldsymbol{x}^{\rm{N}}_{j}\}^{n_{\rm{N}}}_{j=1}$, respectively. Then Eq.~\ref{eq1} can be estimated by $\hat{\mathrm{R}}_{\rm PN}(f) = \pi \hat{\mathrm{R}}_{\rm P}^{+}(f) + (1 - \pi) \hat{\mathrm{R}}_{\rm N}^{-}(f)$, where $\hat{\mathrm{R}}_{\rm P}^{+}(f) = \frac{1}{n_{\rm{P}}} \sum_{j=1}^{n_{\rm{P}}} \ell (f(\boldsymbol{x}_{j}^{\rm{P}}), 1)$ and $\hat{\mathrm{R}}_{\rm N}^{-}(f) = \frac{1}{n_{\rm{N}}} \sum_{j = 1}^{n_{\rm{N}}} \ell (f(\boldsymbol{x}_{j}^{\rm{N}}), 0).$

\subsection{Binary PU Learning}
In PU learning setting, we have only access to labeled positive data $\mathcal{X}_{\rm{P}}$ and unlabeled data $\mathcal{X}_{\rm{U}} = \{\boldsymbol{x}^{\rm{U}}_{j}\}^{n_{\rm{U}}}_{j=1}$ drawn from $p_{\rm{U}}(\boldsymbol{x})$ instead of labeled negative data $\mathcal{X}_{\rm{N}}$, which indicates that the classification risk Eq. \ref{eq1} can not be directly estimated as done in supervised learning setting. For this problem, \citet{du2014analysis} propose the expected classification risk formulation of PU learning:
\vspace{-0.5mm}
\begin{align} \label{eq2}
        \mathrm{R}(f) = {} & \pi \mathrm{R}_{\rm{P}}^{+}(f) + \mathrm{R}_{\rm{U}}^{-}(f) - \pi \mathrm{R}_{\rm{P}}^{-}(f), 
        \vspace{-0.5mm}
\end{align}
where $\mathrm{R}_{\rm{U}}^{-}(f) = \mathbb{E}_{\boldsymbol{x} \sim p(\boldsymbol{x})} \left[\ell (f(\boldsymbol{x}), 0)\right]$ and $\mathrm{R}_{\rm{P}}^{-}(f) = \mathbb{E}_{\boldsymbol{x}\sim p(\boldsymbol{x}\mid y=1)} \left[\ell (f(\boldsymbol{x}), 0)\right]$. Here $\mathrm{R}_{\rm{U}}^{-}(f) - \pi \mathrm{R}_{\rm{P}}^{-}(f)$ can alternatively represent $(1 - \pi)\mathrm{R}_{\rm{N}}^{-}(f)$ because $p(y = 0)p(\boldsymbol{x}\mid y = 0) = p(\boldsymbol{x}) - p(y = 1)p(\boldsymbol{x}\mid y = 1)$.

PU learning assumes that unlabeled data $\mathcal{X}_{\rm{U}}$ can reflect the true overall distribution, that is, $p_{\rm{U}}(\boldsymbol{x}) = p(\boldsymbol{x})$, due to unlabeled data consisting of both positive and negative data, under which Eq.~\ref{eq2} can be approximated by $\hat{\mathrm{R}}_{\rm PU}(f) = \pi \hat{\mathrm{R}}_{\rm{P}}^{+}(f) + \hat{\mathrm{R}}_{\rm{U}}^{-}(f) - \pi \hat{\mathrm{R}}_{\rm{P}}^{-}(f)$, where $\hat{\mathrm{R}}_{\rm{U}}^{-}(f) = \frac{1}{n_{\rm{U}}} \sum_{j = 1}^{n_{\rm{U}}} \ell (f(\boldsymbol{x}_{j}^{\rm{U}}), 0)$ and $\hat{\mathrm{R}}_{\rm{P}}^{-}(f) = \frac{1}{n_{\rm{P}}} \sum_{j=1}^{n_{\rm{P}}} \ell (f(\boldsymbol{x}_{j}^{\rm{P}}),0)$.

\section{Methodology} \label{methods}
In this section, we introduce the proposed Conf-MPU learning for DS-NER in the multi-class classification setting with solid theoretical analysis.

\subsection{Conf-MPU Learning} \label{Conf-MPU}
Let $y \in \mathcal{Y} = \{0, 1, 2, ..., k\}$, where $0$ refers to the negative class and $1, ..., k$ refer to $k$ positive classes. The goal in multi-class classification is to minimize the following expected classification risk:
\vspace{-2mm}
\begin{equation} \label{eq3}
        \mathrm{R}(f) = \sum_{i = 1}^{k} \pi_i \mathrm{R}_{{\rm{P}}_i}^{+}(f) + (1 - \sum_{i = 1}^{k} \pi_i) \mathrm{R}_{\rm N}^{-}(f),
        \vspace{-2mm}
\end{equation}
where $\mathrm{R}^{+}_{{\rm P}_i}(f) = \mathbb{E}_{\boldsymbol{x}\sim p(\boldsymbol{x}\mid y=i)} \left[\ell (f(\boldsymbol{x}), i)\right]$ and $\pi_i=p(y = i)$ are the classification risk and the prior of the $i$-th positive class, respectively. We denote this classification risk as MPN.

Following PU learning setting, there are only labeled positive data $\mathcal{X}_{{\rm{P}}_{i}} = \{\boldsymbol{x}^{{\rm{P}}_i}_{j}\}^{n_{{\rm{P}}_i}}_{j=1}$ drawn from $p_{{\rm{P}}_i}(\boldsymbol{x}) = p(\boldsymbol{x} \mid y = i)$ where $i \in \{1, ..., k\}$, and unlabeled data $\mathcal{X}_{\rm{U}}$. Thus we can not directly estimate Eq.~\ref{eq3}. Here we adopt the same probability principle as applied in binary PU learning to alternatively compute the risk on negative data. Since $p(y = 0)p(\boldsymbol{x}\mid y = 0) = p(\boldsymbol{x}) - \sum_{i = 1}^{k} p(y = i)p(\boldsymbol{x}\mid y = i)$, we can further derive Eq.~\ref{eq3} as:
\vspace{-2mm}
\begingroup\makeatletter\def\f@size{9}\check@mathfonts
\def\maketag@@@#1{\hbox{\m@th\large\normalfont#1}}%
\begin{equation} \label{eq4}
      \mathrm{R}(f) = \sum_{i = 1}^{k} \pi_i \mathrm{R}_{{\rm{P}}_i}^{+}(f) + \mathrm{R}_{\rm{U}}^{-}(f) - \sum_{i = 1}^{k} \pi_i \mathrm{R}_{{\rm{P}}_i}^{-}(f), 
      \vspace{-2mm}
\end{equation}\endgroup
where $\small{\mathrm{R}_{\rm{U}}^{-}(f) - \sum_{i = 1}^{k} \pi_i \mathrm{R}_{{\rm{P}}_i}^{-}(f)}$ theoretically plays the role of $(1 - \sum_{i = 1}^{k} \pi_i) \mathrm{R}_{\rm N}^{-}(f)$, and $\mathrm{R}_{{\rm{P}}_i}^{-}(f) = \mathbb{E}_{\boldsymbol{x}\sim p_{{\rm{P}}_i}(\boldsymbol{x})} \left[\ell (f(\boldsymbol{x}), 0)\right]$. We denote this classification risk as MPU, whose estimation requires the same assumption of data distribution as binary PU learning does, namely, $p_{\rm{U}}(\boldsymbol{x}) = p(\boldsymbol{x})$. We refer this assumption to \textbf{PU assumption} hereinafter for convenience. Under PU assumption, $\mathcal{X}_{\rm{U}}$ can be used to estimate $\mathrm{R}_{\rm{U}}^{-}(f)$. Specifically, MPU risk estimator is given as:
\vspace{-2mm}
\begingroup\makeatletter\def\f@size{8}\check@mathfonts
\def\maketag@@@#1{\hbox{\m@th\large\normalfont#1}}%
\begin{align} \label{eq5}
    &\hat{\mathrm{R}}_{\rm MPU}(f) = \sum_{i=1}^{k} \frac{\pi_i}{n_{{\rm{P}}_i}} \sum_{j=1}^{n_{{\rm{P}}_i}} \ell (f(\boldsymbol{x}_{j}^{{\rm{P}}_i}), i) + {\rm{max}}\bigg\{0, \nonumber \\
    &\quad \frac{1}{n_{\rm{U}}} \sum_{j = 1}^{n_{\rm{U}}}\ell (f(\boldsymbol{x}_{j}^{\rm{U}}), 0) - \sum_{i=1}^{k} \frac{\pi_i}{n_{{\rm{P}}_i}} \sum_{j=1}^{n_{{\rm{P}}_i}} \ell (f(\boldsymbol{x}_{j}^{{\rm{P}}_i}), 0)\bigg\},
    \vspace{-2mm}
\end{align}\endgroup
with a non-negative constraint inspired by \citet{kiryo2017positive} ensuring the risk on the negative class is non-negative.

However, PU assumption can be violated in real distant supervision scenarios. The distribution of unlabeled data $p_{\rm{U}}(\boldsymbol{x})$ may be different from the overall distribution $p(\boldsymbol{x})$ especially when the distant supervision has a good coverage. In such cases, the unlabeled data will have a distribution closer to the distribution of the true negative data $p_{\rm{N}}(\boldsymbol{x})$ instead of the overall distribution $p(\boldsymbol{x})$.
Thus, the risk estimation of $\mathrm{R}_{\rm{U}}^{-}(f)$ based on the assumption, in either MPU or binary PU, may be biased. 

To alleviate such estimation bias, we derive a novel Conf-MPU risk function from MPU. With the context of NER tasks, we observe that almost any combination of characters could be part of a named entity. 
Based on this observation, mathematically, we define $\lambda(\boldsymbol{x}) = p(y > 0 \mid \boldsymbol{x})$ to determine the confidence score of a token being an entity token, no matter what entity type it belongs to, and further assume that $\lambda(\boldsymbol{x}) > 0$. Under this assumption, we can further decompose $\mathrm{R}_{\rm{U}}^{-}(f)$ in Eq.~\ref{eq4} by involving a threshold parameter $0 < \tau \leq 1$ as follows:
\vspace{-2mm}
\begingroup\makeatletter\def\f@size{10}\check@mathfonts
\def\maketag@@@#1{\hbox{\m@th\large\normalfont#1}}%
\begin{equation} \label{eq6}
       \mathrm{R}_{\rm{U}}^{-}(f) = \sum_{i=1}^{k} \pi_i \mathrm{R}_{\thicktilde{\rm{P}}_i}^{-}(f) + \mathrm{R}_{\thicktilde{\rm{U}}}^{-}(f),
      \vspace{-2mm}
\end{equation}\endgroup
where $\small{\mathrm{R}_{\thicktilde{\rm{P}}_i}^{-}(f) = \mathbb{E}_{\boldsymbol{x} \sim p_{{\rm{P}}_i}(\boldsymbol{x} \mid \lambda(\boldsymbol{x}) > \tau)}\left[\ell (f(\boldsymbol{x}), 0) \frac{1}{\lambda (\boldsymbol{x})}\right]}$ and $\mathrm{R}_{\thicktilde{\rm{U}}}^{-}(f) = \mathbb{E}_{\boldsymbol{x} \sim p(\boldsymbol{x} \mid \lambda(\boldsymbol{x}) \leq \tau)}\left[\ell (f(\boldsymbol{x}), 0) \right]$. The detailed proof is shown as follows.\\
\noindent \textbf{Proof.} Since $\lambda(\boldsymbol{x}) > 0$ and $0 < \tau \leq 1$, we have
\vspace{-2mm}
\begingroup\makeatletter\def\f@size{8}\check@mathfonts
\def\maketag@@@#1{\hbox{\m@th\large\normalfont#1}}%
\begin{align} 
    &\mathrm{R}_{\rm{U}}^{-}(f) = \mathbb{E}_{\boldsymbol{x} \sim p(\boldsymbol{x})} \left[\ell (f(\boldsymbol{x}), 0)\right]\nonumber\\
    & = \int_{\lambda(\boldsymbol{x}) > \tau} \ell (f(\boldsymbol{x}), 0) p(\boldsymbol{x})d\boldsymbol{x}
    + \int_{\lambda(\boldsymbol{x}) \leq \tau} \ell (f(\boldsymbol{x}), 0) p(\boldsymbol{x})d\boldsymbol{x}\nonumber\\
    & = \int_{\lambda(\boldsymbol{x}) > \tau} \ell (f(\boldsymbol{x}), 0) \frac{p(\boldsymbol{x})p(\boldsymbol{x}, y>0)}{p(\boldsymbol{x}, y>0)}d\boldsymbol{x} + \mathrm{R}_{\thicktilde{\rm{U}}}^{-}(f) \nonumber\\
    & = \int_{\lambda(\boldsymbol{x}) > \tau} \ell (f(\boldsymbol{x}), 0) \frac{p(\boldsymbol{x})p(\boldsymbol{x} \mid y>0)p(y>0)}{p(y > 0 \mid \boldsymbol{x})p(\boldsymbol{x})}d\boldsymbol{x} + \mathrm{R}_{\thicktilde{\rm{U}}}^{-}(f) \nonumber\\
    & = \int_{\lambda(\boldsymbol{x}) > \tau} \ell (f(\boldsymbol{x}), 0) \frac{p(\boldsymbol{x} \mid y>0)p(y>0)}{ \lambda(\boldsymbol{x})}d\boldsymbol{x} + \mathrm{R}_{\thicktilde{\rm{U}}}^{-}(f) \nonumber\\
    & = \sum_{i = 1}^{k} p(y = i) \int_{\lambda(\boldsymbol{x}) > \tau} \ell (f(\boldsymbol{x}), 0) \frac{p(\boldsymbol{x} \mid y=i)}{\lambda(\boldsymbol{x})}d\boldsymbol{x} + \mathrm{R}_{\thicktilde{\rm{U}}}^{-}(f) \nonumber\\
    & = \sum_{i = 1}^{k} \pi_{i} \int_{\lambda(\boldsymbol{x}) > \tau} \ell (f(\boldsymbol{x}), 0) \frac{1}{\lambda(\boldsymbol{x})}p(\boldsymbol{x} \mid y=i)d\boldsymbol{x} + \mathrm{R}_{\thicktilde{\rm{U}}}^{-}(f) \nonumber\\
    & = \sum_{i=1}^{k} \pi_i \mathrm{R}_{\thicktilde{\rm{P}}_i}^{-}(f) + \mathrm{R}_{\thicktilde{\rm{U}}}^{-}(f). \nonumber
    \vspace{-2mm}
\end{align} \endgroup

Consequently, we obtain the expected classification risk of Conf-MPU by substituting $\mathrm{R}_{\rm{U}}^{-}(f)$ in Eq.~\ref{eq4} with Eq.~\ref{eq6} as follows:
\vspace{-2mm}
\begingroup\makeatletter\def\f@size{8.5}\check@mathfonts
\def\maketag@@@#1{\hbox{\m@th\large\normalfont#1}}%
\begin{equation} \label{eq7}
        \mathrm{R}(f) = \sum_{i=1}^{k}\pi_i \big(\mathrm{R}_{{\rm{P}}_i}^{+}(f) + \mathrm{R}_{\thicktilde{\rm{P}}_i}^{-}(f) - \mathrm{R}_{{\rm{P}}_i}^{-}(f)\big) + \mathrm{R}_{\thicktilde{\rm{U}}}^{-}(f).
        \vspace{-2mm}
\end{equation}\endgroup

Given a reliable $\lambda$ and a proper $\tau$, $\lambda(\boldsymbol{x}) > \tau$ indicates $\boldsymbol{x}$ being an entity token (a positive sample), otherwise a non-entity token (a negative sample), which further induces that $p_{{\rm{P}}_i}(\boldsymbol{x} \mid \lambda(\boldsymbol{x}) > \tau) \approx p_{{\rm{P}}_i}(\boldsymbol{x})$, and $p(\boldsymbol{x} \mid \lambda(\boldsymbol{x}) \leq \tau) \approx p_{\rm{U}}(\boldsymbol{x} \mid \lambda(\boldsymbol{x}) \leq \tau) \approx p_{\rm{N}}(\boldsymbol{x})$ even if $p_{\rm{U}}(\boldsymbol{x})$ is different from $p(\boldsymbol{x})$. Thus, empirically $\mathrm{R}_{\thicktilde{\rm{P}}_i}^{-}(f)$ and $\mathrm{R}_{\thicktilde{\rm{U}}}^{-}(f)$ can be estimated with less bias using $\mathcal{X}_{{\rm{P}}_i}$ and $\mathcal{X}_{\rm{U}}$, respectively, which further leads to a more precise estimation of $\mathrm{R}_{\rm{U}}^{-}(f)$. This is the mechanism that Conf-MPU can significantly reduce estimation bias in practice, even if PU assumption is violated.
Specifically, Conf-MPU risk estimator can be expressed as:
\vspace{-1mm}
\begingroup\makeatletter\def\f@size{9}\check@mathfonts
\def\maketag@@@#1{\hbox{\m@th\large\normalfont#1}}%
\begin{align} \label{eq8}
    &\hat{\mathrm{R}}_{\rm Conf-MPU}(f) = \sum_{i = 1}^{k} \frac{\pi_i}{n_{{\rm{P}}_i}} \sum_{j=1}^{n_{{\rm{P}}_i}} {\rm max}\bigg\{0, \ell (f(\boldsymbol{x}_{j}^{{\rm{P}}_i}), i) \nonumber \\
    & \quad + \mathds{1}_{\hat{\lambda}(\boldsymbol{x}_{j}^{{\rm{P}}_i}) > \tau} \ell (f(\boldsymbol{x}_{j}^{{\rm{P}}_i}), 0)\frac{1}{\hat{\lambda}(\boldsymbol{x}_{j}^{{\rm{P}}_i})} - \ell (f(\boldsymbol{x}_{j}^{{\rm{P}}_i}), 0) \bigg\} \nonumber\\
    & \quad + \frac{1}{n_{\rm{U}}} \sum_{j = 1}^{n_{\rm{U}}} \left[\mathds{1}_{\hat{\lambda}(\boldsymbol{x}_{j}^{\rm U}) \leq \tau} \ell (f(\boldsymbol{x}_{j}^{\rm{U}}), 0)\right],
\end{align}\endgroup
with a constraint to guarantee a non-negative loss on each labeled positive sample,
where $\hat{\lambda}$ is an empirical confidence score estimator. In DS-NER tasks, we formulate the sub-task of estimating $\lambda(\boldsymbol{x})$ as a token-level binary classification problem which also uses distant labels. In practice, a classifier with a sigmoid output layer for this sub-task can guarantee $\hat{\lambda}(\boldsymbol{x}) > 0$.

\subsection{Insights into Conf-MPU Risk Estimator} \label{insights} 
Targeting on the challenge of high false negative rates in training data, we give the following analysis to offer some insights into the Conf-MPU risk estimator. 
For ease of expression, we use letters to denote the terms in Eq.~(\ref{eq8}):
    $\mathbb{A}$ = $\ell (f(\boldsymbol{x}_{j}^{{\rm{P}}_i}), i)$,
    $\mathbb{B}$ = $\mathds{1}_{\hat{\lambda}(\boldsymbol{x}_{j}^{{\rm{P}}_i}) > \tau} \ell (f(\boldsymbol{x}_{j}^{{\rm{P}}_i}), 0)\frac{1}{\hat{\lambda}(\boldsymbol{x}_{j}^{{\rm{P}}_i})}$,
    $\mathbb{C}$ = $\ell (f(\boldsymbol{x}_{j}^{{\rm{P}}_i}), 0)$,
    $\mathbb{D}$ = $\mathds{1}_{\hat{\lambda}(\boldsymbol{x}_{j}^{\rm U}) \leq \tau} \ell (f(\boldsymbol{x}_{j}^{\rm{U}}), 0)$.
The threshold $\tau$ is set to 0.5 by default. We assume that $\hat{\lambda}(\boldsymbol{x})$ of an entity token is close to 1 (\textit{i.e.}, $\hat{\lambda}(\boldsymbol{x}) > \tau$), and $\hat{\lambda}(\boldsymbol{x})$ of a non-entity token is close to 0 (\textit{i.e.}, $\hat{\lambda}(\boldsymbol{x}) \leq \tau$). 

For a true positive sample (\textit{e.g.}, \say{sepsis} distantly labeled in Figure \ref{fig:figure_1}), the loss is computed by $\mathbb{A} + \mathbb{B} - \mathbb{C}$, where $\mathbb{B}$ is involved because its confidence score is larger than the threshold. Since $1/\hat{\lambda}(\boldsymbol{x})$ is close to 1, $\mathbb{B} - \mathbb{C}$ is almost 0 but positive, and thus the loss on this sample approximately equals to $\mathbb{A}$, which is very similar with the loss on a positive sample in standard supervised learning. For a true negative sample (\textit{e.g.}, \say{patient} unlabeled in Figure \ref{fig:figure_1}), the loss is calculated by $\mathbb{D}$ due to its confidence score is less than the threshold. So the minimization for $\mathbb{D}$ enables the model to learn from this true negative sample. For a false negative sample (\textit{e.g.}, \say{neutropenia} unlabeled in Figure \ref{fig:figure_1}), the loss is not counted, because its confidence score is larger than the threshold and thus $\mathbb{D}$ is not calculated. It is the mechanism that Conf-MPU handles false negative samples from unlabeled data.


\subsection{Estimation Error Bound}
Here we establish an estimation error bound for the proposed Conf-MPU risk estimator (Eq.~\ref{eq8}) to show the guaranteed performance. 
\begin{customthm}{1}\label{theorem 1}
Let $f^* = \argmin_{f \in \mathcal{F}}\mathrm{R}(f)$ and $\hat{f}_{\rm Conf-MPU} = \argmin_{f \in \mathcal{F}} \hat{\mathrm{R}}_{\rm Conf-MPU}(f)$. Assume that $\ell(\cdot) \in [0, C_l]$ and $\ell$ is Lipschitz continuous on the interval $[-C_g, C_g]$ with a Lipschitz constant $L_l$, where $C_l, C_g > 0$. Also suppose that $\hat{\lambda}$ is a fixed function independent of data used to compute $\hat{\mathrm{R}}_{\rm Conf-MPU}(f)$ and $\tau \in (0, 1]$. Let $\zeta = p(\hat{\lambda}(\boldsymbol{x}) \leq \tau)$ and $\epsilon = \mathbb{E}_{\boldsymbol{x} \sim p(\boldsymbol{x})} \left[|\hat{\lambda}(\boldsymbol{x}) - \lambda(\boldsymbol{x})|^2\right]$. Then for any $\delta > 0$, with probability at least $1 - \delta$,
\vspace{-2mm}
\begingroup\makeatletter\def\f@size{8.5}\check@mathfonts
\def\maketag@@@#1{\hbox{\m@th\large\normalfont#1}}%
\begin{align*}
    & \mathrm{R}(\hat{f}_{\rm Conf-MPU}) - \mathrm{R}(f^*) \leq \sum_{i = 1}^{k} 2\pi_i \frac{(\tau + 1)C_l}{\tau}\\
    & + \sum_{i = 1}^{k}2\pi_i \left[ \frac{2L_l}{\tau}\mathfrak{R}_{n_{{\rm{P}}_i}, p_{{\rm{P}}_i}(\boldsymbol{x})}(\mathcal{F}) 
     + \frac{(\tau + 1)C_l}{\tau}\sqrt{\frac{\log \frac{k + 1}{\delta}}{2n_{{\rm{P}}_i}}}\right] \\
     & + 4 L_l \mathfrak{R}_{n_{\rm{U}}, p(\boldsymbol{x})}(\mathcal{F}) + 2C_l\sqrt{\frac{\log \frac{k + 1}{\delta}}{2n_{\rm U}}}
     + \frac{2C_l}{\tau}\sqrt{(1 - \zeta)\epsilon}.
\end{align*}\endgroup
\end{customthm}
In Theorem \ref{theorem 1}, $\mathcal{F}$ is the function class and $\mathfrak{R}_{n_{{\rm{P}}_i}, p_{{\rm{P}}_i}(\boldsymbol{x})}(\mathcal{F})$ is the Rademacher complexity of the function class $\mathcal{F}$ for the sampling of size $n_{{\rm{P}}_i}$ from the distribution $p_{{\rm{P}}_i}(\boldsymbol{x})$ and $\mathfrak{R}_{n_{\rm{U}}, p(\boldsymbol{x})}(\mathcal{F})$ follows a similar definition. We relegate this proof to the Appendix.

\section{DS-NER Classification}
In this section, we describe the setup for the DS-NER classification.

\subsection{Generation of Distant Labels}
In DS-NER tasks, professional dictionaries (\textit{e.g.}, UMLS) and knowledge bases (\textit{e.g.}, Wikidata) are used to automatically generate distant labels. Distant labeling by dictionaries employs some string matching algorithms to map training samples to dictionaries \cite{ren2015clustype, giannakopoulos2017unsupervised, peng2019distantly}, while knowledge bases utilize public APIs to perform such distant labeling.

\subsection{Classifiers} \label{classifier}
The proposed Conf-MPU risk estimator can be applied on any NER classifiers where the task is to predict the label for each token. For example, BERT \cite{devlin-etal-2019-bert} can be used as the underlying NER model, and then the Conf-MPU risk estimation can be used to calculate the classification risks. We use \textbf{Conf-MPU{\scriptsize\texttt{BERT}}} to denote this method. 
BiLSTM \cite{chiu2016named} is another popular choice for NER models. 
\citet{ratinov2009design, passos2014lexicon, chiu2016named} demonstrate that using lexicons as external features can improve NER performance. With the dictionaries, we extract the lexicon features as follows. For each token, we match its contextual words within a window size against entries in the dictionaries. If there is any successful matching, a binary indicator is set to 1, otherwise 0. With the window size of $n$, we can form an $n$-bit vector, which is appended to the input embedding. We denote the model of BiLSTM with the lexicon feature engineering as LBiLSTM, and denote \textbf{Conf-MPU{\scriptsize\texttt{LBiLSTM}}} as the LBiLSTM-based classifier with Conf-MPU risk estimation.


For the first step of estimating confidence scores, we build a token-level binary classifier (\textit{i.e.}, $\lambda$) based on LBiLSTM to output scores. To be consistent with PU learning setting, this classifier is equipped with a binary PU learning risk estimator (\textit{i.e.}, $\hat{\mathrm{R}}_{\rm PU}(\lambda)$).

\subsection{Prior Estimation}\label{sec:prior}
Unlike in supervised learning where priors (\textit{i.e.}, $\pi_i$) can be easily obtained from human annotations, we cannot directly acquire them from distant annotations. In PU learning research, there are some methods proposed specifically for estimating the priors \cite{bekker2018estimating, jain2016estimating, du2014class}. Here we adopt the most effective TIcE algorithm from \citet{bekker2018estimating} to perform prior estimation.

\subsection{Loss Function}
\citet{peng2019distantly} point out that a bounded loss function can help avoid overfitting in PU learning setting. We also confirm this argument in our empirical studies. Thus, instead of using the common unbounded cross entropy loss function, we adopt the mean absolute error (MAE) as the loss function for Conf-MPU and other PU learning methods in our experiments. Given its label $\boldsymbol{y}$ in the one-hot form, the loss on a token $\boldsymbol{x}$ is defined by:
\vspace{-2mm}
\begin{equation*}
    \ell (f(\boldsymbol{x}), \boldsymbol{y}) = \frac{1}{k + 1} \sum_{i=0}^{k} |\boldsymbol{y}^{(i)} - f(\boldsymbol{x})^{(i)}|,
    \vspace{-1mm}
\end{equation*}
where $f(\boldsymbol{x})$ is the softmax output, and both $\boldsymbol{y}$ and $f(\boldsymbol{x})$ are in $k + 1$ dimensions. Note that $\ell (f(\boldsymbol{x}), \boldsymbol{y}) \in [0, \frac{2}{k + 1}]$ is bounded. 

\subsection{Post-Processing}
In DS-NER tasks, self-training strategies as post-processing can often further improve the performance, such as iteratively enriching dictionaries based on the model predictions \cite{peng2019distantly}, or iteratively training a teacher-student framework \cite{liang2020bond}. The discussion for self-training framework is out of the scope of this paper and we refer the readers to \citet{zoph2020rethinking} for more information.  

\section{Experiments} \label{Experiments}
In this section, we evaluate the proposed Conf-MPU and compare with other baseline methods.

\subsection{Experimental Setup}
\subsubsection{Training Data and Evaluation Metrics}\label{datasets}
We consider two benchmark NER datasets from different domains: (1) \textbf{BC5CDR} comes from biomedical domain. It consists of 1,500 articles, containing 15,935 \texttt{Chemical} and 12,852 \texttt{Disease} mentions; (2) \textbf{CoNLL2003} is a well-known open-domain NER dataset. It consists of 1,393 English news articles, containing 10,059 \texttt{PER}, 10,645 \texttt{LOC}, 9,323 \texttt{ORG} and 5,062 \texttt{MISC} mentions. 

We obtain the following distantly labeled datasets: (1) \textbf{BC5CDR (Big Dict)} is labeled using a dictionary\footnote{\url{https://github.com/shangjingbo1226/AutoNER}} released by \citet{shang2018learning}; (2) \textbf{BC5CDR (Small Dict)} is labeled using a smaller dictionary constructed by selecting only the first 20\% entries from the previous one; (3) \textbf{CoNLL2003 (KB)}\footnote{\url{https://github.com/cliang1453/BOND}} is labeled by the knowledge base Wikidata and released by \citet{liang2020bond}; (4) \textbf{CoNLL2003 (Dict)} is labeled using a refined dictionary released by \citet{peng2019distantly}\footnote{\url{https://github.com/v-mipeng/LexiconNER}}. For dictionary labeling, we use the strict string matching algorithm presented in \citet{peng2019distantly}. The process of knowledge base labeling can be found in \citet{liang2020bond}. 

All DS-NER methods are trained on the same distantly labeled training data and evaluated on the released human-annotated test sets in terms of span-level precision, recall and F1 score.

\subsubsection{Baseline Methods} 
We compare the proposed Conf-MPU with different groups of baseline methods.

\paragraph{Fully Supervised Methods.} We present the state-of-the-art (SOTA) performance of fully supervised methods on the two benchmark datasets, \citet{wang2021improving} on BC5CDR and \citet{wang2020automated} on CoNLL2003. For SOTA methods, we report the results from their original papers. We also evaluate the employed BiLSTM and BERT models in fully supervised setting. The performance in this group serves as upper-bound references.

\paragraph{Distantly Supervised Methods.} We consider the following distantly supervised NER methods: (1) \textbf{Dict/KB Matching} distantly labels the test sets using dictionaries or knowledge bases directly, which is included here as references; (2) \textbf{AutoNER} \cite{shang2018learning} trains the model using a ``tie-or-break'' mechanism to detect entity boundaries and then predicts entity type for each candidate; (3) \textbf{BERT-ES} \cite{liang2020bond} adopts early stopping to prevent BERT from overfitting to noisy distant labels; 
(4) \textbf{BNPU} \cite{peng2019distantly} built on LBiLSTM (\textbf{BNPU{\scriptsize\texttt{LBiLSTM}}}) applies a binary PU learning risk estimation with MAE as the loss function to each entity type and then infers the final types; (5) \textbf{MPU} is the predecessor of the proposed Conf-MPU, which computes the empirical risk using Eq.~\ref{eq5}. We also build MPU on both BERT and LBiLSTM models, denoted as \textbf{MPU{\scriptsize\texttt{BERT}}} and \textbf{MPU{\scriptsize\texttt{LBiLSTM}}}. 
Note that full models in \citet{peng2019distantly, liang2020bond} contain self-training as post processing steps, which are omitted here. We focus on the evaluation of how well each model can handle incomplete labeling issues in DS-NER tasks.

\subsubsection{Method Setups}
To evaluate the efficacy of the DS-NER methods in real usage under distantly supervised settings, we do not use any human-annotated validation or test sets in any stage of the training process. The training stopping criteria are set as follows: 100 epochs for BiLSTM-based methods and 5 epochs for BERT-based ones. We report the performance of the final model instead of the best checkpoint. Consequently, the baselines have different performance from their reported results.
We use the released code for AutoNER and BERT-ES to reproduce their results. For other methods, we report the results based on our implementations. BiLSTM-based models utilize pretrained \texttt{bio-embedding}\footnote{{\tiny \url{https://github.com/shangjingbo1226/AutoNER}}} for BC5CDR and pretrained Stanford's \texttt{Glove}\footnote{\url{https://nlp.stanford.edu/projects/glove/}} embedding for CoNLL2003. BERT-based models use pretrained \texttt{biobert-base-cased-v1.1}\footnote{\url{https://huggingface.co/dmis-lab/biobert-base-cased-v1.1}} for BC5CDR and \texttt{bert-base-cased}\footnote{\url{https://huggingface.co/bert-base-cased}} for CoNLL2003. The only exception is that BERT-ES uses \texttt{roberta-base}\footnote{\url{https://huggingface.co/roberta-base}} for CoNLL2003 in the original implementation.



\begin{table}[t]
\centering
\resizebox{0.45\textwidth}{!}{%
\begin{tabular}{llll}
\Xhline{3\arrayrulewidth}
Dataset                                       & Type     & Precision & Recall \\
\Xhline{2\arrayrulewidth}
\multirow{2}{*}{\textbf{BC5CDR (Big Dict)}}   & Chemical & 97.99     & 63.14  \\
                                              & Disease  & 98.34     & 46.73  \\
\Xhline{1.5\arrayrulewidth}
\multirow{2}{*}{\textbf{BC5CDR (Small Dict)}} & Chemical & 98.66     & 11.43  \\
                                              & Disease  & 99.25     & 9.31   \\
\Xhline{1.5\arrayrulewidth}
\multirow{4}{*}{\textbf{CoNLL2003 (KB)}}      & PER      & 82.36     & 82.11  \\
                                              & LOC      & 99.98     & 65.20  \\
                                              & ORG      & 90.47     & 60.59  \\
                                              & MISC     & 100.00    & 20.07  \\
\Xhline{1.5\arrayrulewidth}
\multirow{4}{*}{\textbf{CoNLL2003 (Dict)}}    & PER      & 99.78     & 79.10  \\
                                              & LOC      & 97.56     & 34.69  \\
                                              & ORG      & 95.80     & 65.47  \\
                                              & MISC     & 99.24     & 57.22  \\
\Xhline{2\arrayrulewidth}
\end{tabular}%
}
\caption{The quality of distant labels on training sets, stated in token-level precision and recall (in \%)
.}
\vspace{-3mm}
\label{tab:matching}
\end{table}

\begin{table*}[t]
\small
\centering
\resizebox{1\textwidth}{!}{%
\begin{tabular}{lcccc}
\Xhline{3\arrayrulewidth}
Method              &\textbf{BC5CDR (Big Dict)} &\textbf{BC5CDR (Small Dict)} &\textbf{CoNLL2003 (KB)} &\textbf{CoNLL2003 (Dict)} \\
\Xhline{2\arrayrulewidth}
\multicolumn{5}{l}{\textbf{Fully Supervised}}                                                                       \\
\Xhline{2\arrayrulewidth}
Existing SOTA       & \multicolumn{2}{c}{\textbf{90.99 (-/-)}}      & \multicolumn{2}{c}{\textbf{94.60 (-/-)}}      \\
BERT                & \multicolumn{2}{c}{83.88 (79.75/88.46)}       & \multicolumn{2}{c}{89.03 (88.00/90.08)}       \\
BiLSTM              & \multicolumn{2}{c}{75.60 (71.27/80.49)}       & \multicolumn{2}{c}{86.19 (84.06/88.42)}       \\
\Xhline{2\arrayrulewidth}
\multicolumn{5}{l}{\textbf{Distantly Supervised}}                                                                   \\
\Xhline{2\arrayrulewidth}
Dict/KB Matching    & 64.32 (86.39/51.24)   & 15.69 (80.02/8.70)    & 71.40 (81.13/63.75)   & 63.93 (93.12/48.67)   \\
AutoNER             & 79.99 (82.63/77.52)   & 20.66 (81.47/11.83)   & 67.80 (73.10/63.22)   & 61.19 (82.87/48.50)   \\
BERT-ES             & 73.66 (80.43/67.94)   & 17.21 (75.60/9.71)    & 72.15 (81.38/64.80)   & 63.68 (85.77/50.63)   \\
BNPU\scriptsize\texttt{LBiLSTM}     & 59.24 (48.12/77.06)   & 70.21 (64.93/76.43)   & 78.44 (74.38/82.97)   & 76.11 (73.68/78.70)   \\
MPU\scriptsize\texttt{BERT}            & 68.22 (56.50/86.05)   & 73.91 (70.08/78.18)   & 65.75 (58.79/74.58)   & 67.65 (63.63/72.22)   \\
MPU\scriptsize\texttt{LBiLSTM}      & 60.79 (48.28/82.06)   & 73.25 (67.50/80.07)   & 69.13 (59.46/82.54)   & 71.41 (63.41/81.71)   \\
\hline
Conf-MPU\scriptsize\texttt{BERT}       & 77.22 (69.79/86.42)   & 71.85 (81.02/64.54)   & 79.16 (78.58/79.75)   & 81.89 (81.71/82.08)   \\
Conf-MPU\scriptsize\texttt{LBiLSTM} &\textbf{80.07 (76.63/83.82)} &\textbf{76.18 (82.66/70.64)} &\textbf{80.02 (77.39/82.84)} &\textbf{83.34 (85.79/81.02)}   \\
\Xhline{2\arrayrulewidth}
\end{tabular}%
}
\caption{The span-level results on test sets: F1 score (Precision/Recall) (in \%), where the bests are in bold.}
\label{tab:main}
\end{table*}

\begin{figure*}[t]
    \centering
    \includegraphics[width=1\textwidth]{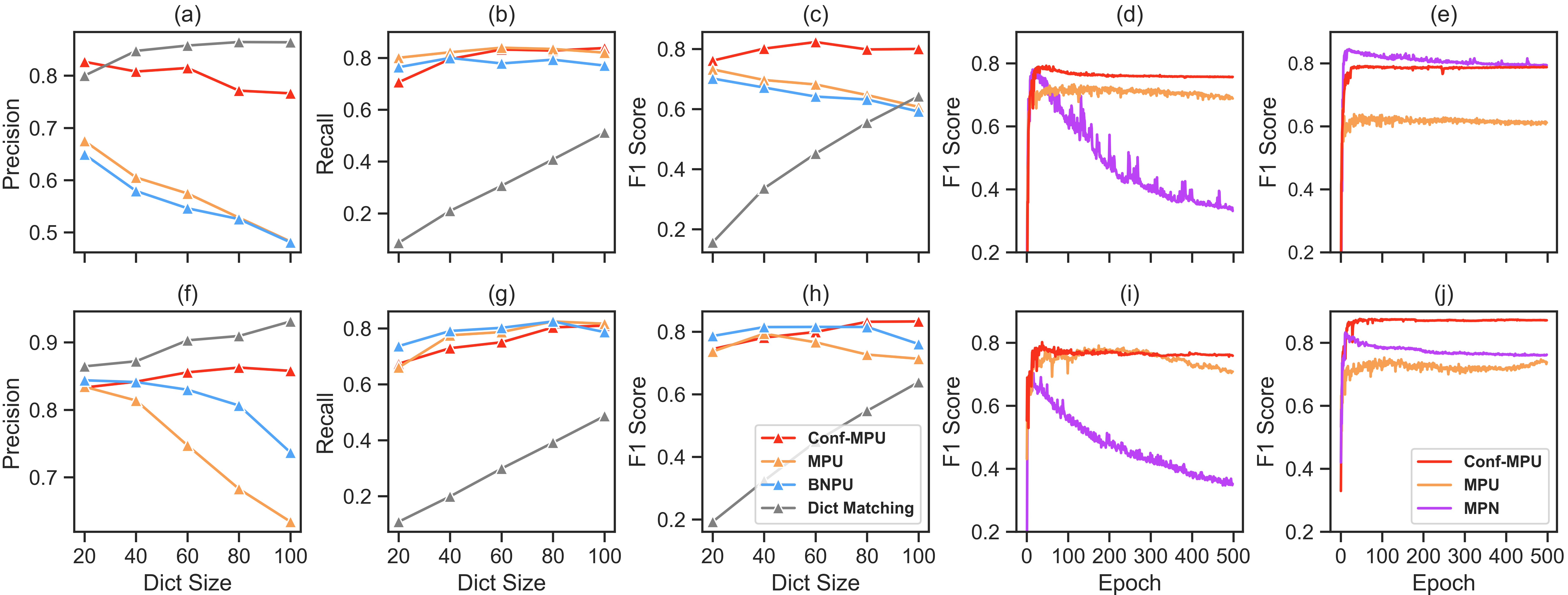}
    \caption{The performance of LBiLSTM-based methods under various settings. Figures in the first row ($\texttt{a}$ - $\texttt{e}$) and the second row ($\texttt{f}$ - $\texttt{j}$) show the results on BC5CDR and CoNLL2003, respectively.}
    \label{fig:2}
    \vspace{-2mm}
\end{figure*}

\subsection{Experimental Results}

\subsubsection{Main Results} \label{general}
We first examine the quality of the distantly labeled training data. 
Table \ref{tab:matching} shows the detailed evaluation of distantly labeled training data. The results validate the assumption mentioned in previous work that distant labels generated by dictionaries are usually of high precision but low recall.

Table \ref{tab:main} presents the overall span-level precision, recall, and F1 scores for all methods on the test sets. 
The proposed Conf-MPU shows a clear advantage over baseline methods, especially when accompanying with LBiLSTM. 
Almost all distantly supervised baselines perform better than Dict/KB Matching on these four datasets, except for a few cases of BNPU and MPU which will be discussed later. Among the baseline methods, AutoNER and BERT-ES show strong correlation with respect to the dictionary quality. On BC5CDR (Small Dict), where the dictionary suffers from extremely low coverage, the two methods have little improvement on recall.

By contrast, all PU learning based methods demonstrate significantly higher recall on all datasets, showing more robustness to the issue of incomplete labeling. However, we can observe that compared with their performances on BC5CDR (Small Dict), BNPU and MPU suffer from low precision on BC5CDR (Big Dict) labeled with a high coverage dictionary. We will extend the discussion in Section \ref{verification}.
As the results manifest, Conf-MPU can significantly improve precision compared with BNPU and MPU, and meanwhile maintain a high level of recall on all datasets, which shows that Conf-MPU can significantly alleviate the estimation bias.
We guide readers to the Appendix for detailed evaluation of prior estimation and confidence score estimation.

\subsubsection{Impact of Dictionary Coverage} \label{verification}
To have a solid recognition to the estimation bias in BNPU and MPU caused by the violation of PU assumption, we construct a series of dictionaries with different coverage on entities. We treat the dictionaries used to generate labels for BC5CDR (Big Dict) and CoNLL2003 (Dict) as two reference standard dictionaries. Then for each of the two benchmark datasets we build a group of dictionaries by selecting the first 20, 40, 60, 80, and 100 (\%) entries from the standard ones. We train BNPU, MPU, and Conf-MPU on the distantly labeled datasets generated by these dictionaries. Here we show the results based on LBiLSTM in Figure \ref{fig:2} ($\texttt{a}$ - $\texttt{c}$ and $\texttt{f}$ - $\texttt{h}$). Similar trend can be observed on BERT-based settings.

We can see a clear decreasing trend on precision for BNPU and MPU when dictionary size increases (Figures \ref{fig:2} ($\texttt{a\&f}$)). These phenomena are caused due to the violation of PU assumption. When a dictionary has higher coverage, the distribution of unlabeled data is more and more similar to the distribution of true negative data, instead of to the overall data distribution. The BNPU and MPU risk estimations bring higher bias, leading to lower precision. Although their recalls remain high, the F1 scores still decrease. By contrast, the proposed Conf-MPU can effectively avoid this limitation and achieve good performance for all dictionary sizes.

\begin{table}[t]
\centering
\resizebox{0.48\textwidth}{!}{%
\begin{tabular}{lcc}
\Xhline{3\arrayrulewidth}
Method   & \textbf{\texttt{LBiLSTM}}               & \textbf{\texttt{BiLSTM}}                \\
\Xhline{2\arrayrulewidth}
BNPU     & 70.21 (64.93/76.43)   & 63.37 (57.92/69.97) \\
MPU      & 73.25 (67.50/80.07)   & 62.39 (56.50/69.66) \\
Conf-MPU & 76.18 (82.66/70.64)   & 68.11 (71.68/64.88) \\
\Xhline{2\arrayrulewidth}
\end{tabular}%
}
\caption{Ablation study on lexicon features on BC5CDR (Small Dict).}
\vspace{-3mm}
\label{tab:ablation}
\end{table}

\subsubsection{Ablation Studies}
To further evaluate the Conf-MPU risk estimation, we first conduct ablation studies comparing with MPN risk (Eq. \ref{eq3}) whose estimation simply treats unlabeled data as negative samples in distant supervision, and demonstrate the performance with different epochs.  Sub-figures ($\texttt{d}$, $\texttt{e}$, $\texttt{i}$, $\texttt{j}$) in Figure \ref{fig:2} show the trends of F1 scores of LBiLSTM-based models on the validation sets using three risk estimations of Conf-MPU, MPU, and MPN, with respect to different number of epochs. 
($\texttt{d}$, $\texttt{e}$) and ($\texttt{i}$, $\texttt{j}$) reflect the performance on BC5CDR and CoNLL2003, respectively. ($\texttt{d}$, $\texttt{i}$) and ($\texttt{e}$, $\texttt{j}$) reflect the performance based on 20\% and 100\% dictionaries, respectively. The results show that MPN risk estimation can lead to severe overfitting for the model when dictionaries have low coverage. Although MPN still causes overfitting on full dictionaries, its performances are more stable and generally good. By contrast, MPU and Conf-MPU consider the false negative issue during training, and do not overfit even on small dictionaries. Conf-MPU performs stably and consistently well for more scenarios comparing with the other risk estimations. 


From Table \ref{tab:main}, we can observe that Conf-MPU{\scriptsize\texttt{LBiLSTM}} outperforms BiLSTM with fully supervised setting on BC5CDR with both big and small dictionaries. To examine the performance gain, we implement three methods, BNPU, MPU, and Conf-MPU based on BiLSTM instead of LBiLSTM to evaluate the impact of lexicon features learned from the dictionaries. The results on BC5CDR (Small Dict) are shown in Table \ref{tab:ablation}. We can see that the lexicon features used in DS-NER tasks can significantly improve the performance. The experiments performed on other distantly labeled datasets also exhibit similar trends. The results suggest that dictionaries in DS-NER tasks can also serve as external features in additional to the distant labels. 

\section{Related Work}
\paragraph{DS-NER.} Here we briefly discuss a few representative approaches. AutoNER \cite{shang2018learning} proposes a new architecture to first determine whether two adjacent tokens should be tied or broken to form entity candidates, and then determine the type for entity candidates. To handle the false negative samples in training data, the detected entity candidates that are unmatched in dictionaries are not counted for the loss calculation during model training. To separate noisy sentences from the training data, \citet{cao2019low} design a data selection scheme to compute scores for annotation confidence and annotation coverage. \citet{mayhew2019named} introduce a constraint driven iterative algorithm learning to detect false negatives in the noisy data and down-weigh them, resulting in a weighted training set. BOND \cite{liang2020bond} leverages the power of pre-trained language model BERT and perform early stopping during training to avoid overfitting to the imperfect annotated labels. 

\paragraph{PU Learning.} PU learning learns a classifier from positive and unlabeled data \cite{elkan2008learning, du2014analysis}. In a broad sense, PU learning belongs to semi-supervised learning. However, there is a fundamental difference between them: semi-supervised learning requires labeled negative data, but PU learning does not. Recently, a few works significantly enriched PU learning theory. \citet{kiryo2017positive} propose a non-negative risk estimator for PU learning, which enables the usage of deep neural networks for classification given limited labeled positive data. \citet{xu2017multi} first come up with the concept of multi-positive and unlabeled learning with a margin maximization goal for the multi-class classification problem. However, the objective of margin maximization cannot be easily extended to apply on popular deep learning architectures. \citet{hsieh2019classification} propose a novel classification framework incorporating biased negative data in PU learning, which opens up a wider range of the applications of PU learning.

\section{Conclusion}
In this paper, we present a novel multi-class positive and unlabeled learning method called Conf-MPU for the DS-NER task. Conf-MPU estimates the empirical classification risks using the confidence estimation on the distantly labeled training data to avoid model overfitting to the false negative samples in unlabeled data. The extensive experiments illustrate that compared with existing DS-NER methods, Conf-MPU is more robust to various types of dictionaries and can handle the incomplete labeling problem effectively.

\endgroup

\bibliography{anthology,custom}
\bibliographystyle{acl_natbib}

\clearpage


\section*{\LARGE Appendix}     
\begingroup\makeatletter\def\f@size{10}\check@mathfonts
\def\maketag@@@#1{\hbox{\m@th\large\normalfont#1}}%

\setcounter{section}{0}
\section{Proof of Theorem \ref{theorem 1}} 
Here we determine the difference between $\mathrm{R}(\hat{f}_{\rm Conf-MPU})$ and $\mathrm{R}(f^*)$ using the error bound between $\hat{\mathrm{R}}_{\rm Conf-MPU}(f)$ and $\mathrm{R}(f)$. Let us first define the intermediate risk estimators as follows. Each one introduces a new estimation component. 

Starting from $\mathrm{R}(f)$ (\textit{i.e.}, Eq.~\ref{eq7}), $\mathrm{\bar{R}}(f)$ introduces $\hat{\lambda}(\boldsymbol{x})$ as the estimation of $\lambda(\boldsymbol{x})$,
\begin{align*}
      \mathrm{\bar{R}}(f) & = \sum_{i=1}^{k} \pi_i \mathbb{E}_{\boldsymbol{x} \sim p(\boldsymbol{x} \mid y = i)} \bigg[ \ell(f(\boldsymbol{x}), i)  \\
       & + \mathds{1}_{\hat{\lambda}(\boldsymbol{x}) > \tau} \ell(f(\boldsymbol{x}), 0)\frac{1}{\hat{\lambda}(\boldsymbol{x})} - \ell(f(\boldsymbol{x}), 0) \bigg]\\
        & + \mathbb{E}_{\boldsymbol{x} \sim p(\boldsymbol{x})} \left[ \mathds{1}_{\hat{\lambda}(\boldsymbol{x}) \leq \tau} \ell(f(\boldsymbol{x}), 0) \right].
\end{align*}

Then, $\hat{\mathrm{R}}(f)$ uses empirical means to estimate expectations,
\begingroup\makeatletter\def\f@size{9.5}\check@mathfonts
\def\maketag@@@#1{\hbox{\m@th\large\normalfont#1}}%
\begin{align*}
      \hat{\mathrm{R}}(f) &= \sum_{i = 1}^{k} \frac{\pi_i}{n_{{\rm{P}}_i}} \sum_{j=1}^{n_{{\rm{P}}_i}} \bigg[\ell (f(\boldsymbol{x}_{j}^{{\rm{P}}_i}), i) \\
    & + \mathds{1}_{\hat{\lambda}(\boldsymbol{x}_{j}^{{\rm{P}}_i}) > \tau} \ell (f(\boldsymbol{x}_{j}^{{\rm{P}}_i}), 0)\frac{1}{\hat{\lambda}(\boldsymbol{x}_{j}^{{\rm{P}}_i})} - \ell (f(\boldsymbol{x}_{j}^{{\rm{P}}_i}), 0)\bigg]\\
    &+ \frac{1}{n_{\rm{U}}} \sum_{j = 1}^{n_{\rm{U}}} \left[\mathds{1}_{\hat{\lambda}(\boldsymbol{x}_{j}^{\rm{U}}) \leq \tau} \ell (f(\boldsymbol{x}_{j}^{\rm{U}}), 0)\right].
\end{align*}\endgroup

$\hat{\mathrm{R}}_{\rm Conf-MPU}(f)$ restricts the loss on each labeled positive sample to be at least 0,
\begingroup\makeatletter\def\f@size{9.5}\check@mathfonts
\def\maketag@@@#1{\hbox{\m@th\large\normalfont#1}}%
\begin{align*}
     & \hat{\mathrm{R}}_{\rm Conf-MPU}(f) = \sum_{i = 1}^{k} \frac{\pi_i}{n_{{\rm{P}}_i}} \sum_{j=1}^{n_{{\rm{P}}_i}} {\rm max} \bigg\{0, \ell (f(\boldsymbol{x}_{j}^{{\rm{P}}_i}), i) \\
    & - \ell (f(\boldsymbol{x}_{j}^{{\rm{P}}_i}), 0) + \mathds{1}_{\hat{\lambda}(\boldsymbol{x}_{j}^{{\rm{P}}_i}) > \tau} \ell (f(\boldsymbol{x}_{j}^{{\rm{P}}_i}), 0)\frac{1}{\hat{\lambda}(\boldsymbol{x}_{j}^{{\rm{P}}_i})}\bigg\} \\
    &+ \frac{1}{n_{\rm{U}}} \sum_{j = 1}^{n_{\rm{U}}} \left[\mathds{1}_{\hat{\lambda}(\boldsymbol{x}_{j}^{\rm{U}}) \leq \tau} \ell (f(\boldsymbol{x}_{j}^{\rm{U}}), 0)\right].
\end{align*}\endgroup

In the following proof, we will derive the error bounds from $\hat{\mathrm{R}}_{\rm Conf-MPU}(f)$ to $\hat{\mathrm{R}}(f)$, from $\hat{\mathrm{R}}(f)$ to $\bar{\mathrm{R}}(f)$, and from $\bar{\mathrm{R}}(f)$ to $\mathrm{R}(f)$ in order.

Let us first derive the error bound from $\hat{\mathrm{R}}_{\rm Conf-MPU}(f)$ to $\hat{\mathrm{R}}(f)$. For ease of notation, let
\begingroup\makeatletter\def\f@size{8.5}\check@mathfonts
\def\maketag@@@#1{\hbox{\m@th\large\normalfont#1}}%
\begin{align*}
     A &= {\rm max}\left\{0, \ell (f(\boldsymbol{x}), i) 
     + \ell (f(\boldsymbol{x}), 0) \left(\mathds{1}_{\hat{\lambda}(\boldsymbol{x}) > \tau} \frac{1}{\hat{\lambda}(\boldsymbol{x})} - 1\right)\right\},\\
     B &= \ell (f(\boldsymbol{x}), i) 
     + \ell (f(\boldsymbol{x}), 0) \left(\mathds{1}_{\hat{\lambda}(\boldsymbol{x}) > \tau} \frac{1}{\hat{\lambda}(\boldsymbol{x})} - 1\right).
\end{align*}\endgroup
Then we have
\begin{align*}
      \left|\hat{\mathrm{R}}_{\rm Conf-MPU}(f) - \hat{\mathrm{R}}(f)\right| &= \sum_{i = 1}^{k} \frac{\pi_i}{n_{{\rm{P}}_i}} \sum_{j=1}^{n_{{\rm{P}}_i}} \left|A - B\right|.
\end{align*}
Since $\ell(\cdot) \in [0, C_l]$, we have $A \in \left[0, \frac{C_l}{\tau}\right]$ and $B \in \left[-C_l, \frac{C_l}{\tau}\right]$. Further, we have $|A - B| \leq \left(1 + \frac{1}{\tau}\right)C_l$. \\
So we get the error bound
\begin{align*}
      \left|\hat{\mathrm{R}}_{\rm Conf-MPU}(f) - \hat{\mathrm{R}}(f)\right| \leq \sum_{i = 1}^{k} \pi_i \frac{(\tau + 1)C_l}{\tau}.
\end{align*}

Then, we use the following lemma to establish the error bound from $\hat{\mathrm{R}}(f)$ to $\bar{\mathrm{R}}(f)$.

\paragraph{Lemma 1.}\emph{Let $\hat{\lambda}(\cdot): \mathbb{R}^d \rightarrow (0, 1]$ be a fixed function independent of data used to compute $\hat{\mathrm{R}}(f)$ and $\tau \in (0, 1]$. For any $\delta > 0$, with probability at least $1 - \delta$,}
\begingroup\makeatletter\def\f@size{9}\check@mathfonts
\def\maketag@@@#1{\hbox{\m@th\large\normalfont#1}}%
\begin{align*}
    & \sup_{f \in \mathcal{F}} \left|\hat{\mathrm{R}}(f) - \bar{\mathrm{R}}(f)\right| \leq \\
    &\sum_{i = 1}^{k}\pi_i \left[ \frac{2L_l}{\tau}\mathfrak{R}_{n_{{\rm{P}}_i}, p(\boldsymbol{x} \mid y = i)}(\mathcal{F}) 
     + \frac{(\tau + 1)C_l}{\tau}\sqrt{\frac{\log \frac{k + 1}{\delta}}{2n_{{\rm{P}}_i}}}\right] \\
     &+ 2 L_l \mathfrak{R}_{n_{\rm{U}}, p(\boldsymbol{x})}(\mathcal{F}) + C_l\sqrt{\frac{\log \frac{k + 1}{\delta}}{2n_{\rm U}}}.
\end{align*}\endgroup
\emph{Proof of Lemma 1.} For ease of notation, let
\begingroup\makeatletter\def\f@size{9}\check@mathfonts
\def\maketag@@@#1{\hbox{\m@th\large\normalfont#1}}%
\begin{align*}
      \bar{\mathrm{R}}_{{\rm{P}}_i}^{+} = {} & \mathbb{E}_{\boldsymbol{x} \sim p(\boldsymbol{x} \mid y = i)} \Bigg[\ell(f(\boldsymbol{x}), i) \\
      & + \ell(f(\boldsymbol{x}), 0)\left(\mathds{1}_{\hat{\lambda}(\boldsymbol{x})}\frac{1}{\hat{\lambda}(\boldsymbol{x})} - 1\right)\Bigg],\\
      \bar{\mathrm{R}}_{\rm U}^{-} = {} & \mathbb{E}_{\boldsymbol{x} \sim p(\boldsymbol{x})} \left[\mathds{1}_{\hat{\lambda}(\boldsymbol{x}) \leq \tau} \ell (f(\boldsymbol{x}, 0))\right],\\
      \hat{\mathrm{R}}_{{\rm{P}}_i}^{+} = {} & \frac{1}{n_{{\rm{P}}_i}} \sum_{j=1}^{n_{{\rm{P}}_i}} \Bigg[\ell (f(\boldsymbol{x}_{j}^{{\rm{P}}_i}), i) \\
     & + \ell (f(\boldsymbol{x}_{j}^{{\rm{P}}_i}), 0) \left(\mathds{1}_{\hat{\lambda}(\boldsymbol{x}_{j}^{{\rm{P}}_i}) > \tau} \frac{1}{\hat{\lambda}(\boldsymbol{x}_{j}^{{\rm{P}}_i})} - 1\right)\Bigg],\\
     \hat{\mathrm{R}}_{\rm U}^{-} = {} & \frac{1}{n_{\rm{U}}} \sum_{j = 1}^{n_{\rm{U}}} \left[\mathds{1}_{\hat{\lambda}(\boldsymbol{x}_{j}^{\rm{U}}) \leq \tau} \ell (f(\boldsymbol{x}_{j}^{\rm{U}}), 0)\right].
\end{align*}\endgroup

From the sub-additivity of the supremum operator, we have
\begin{align*}
    \sup_{f \in \mathcal{F}} \left|\hat{\mathrm{R}}(f) - \bar{\mathrm{R}}(f)\right| \leq {} & \sum_{i=1}^{k} \pi_i \sup_{f \in \mathcal{F}} \left|\hat{\mathrm{R}}_{{\rm{P}}_i}^{+} - \bar{\mathrm{R}}_{{\rm{P}}_i}^{+}\right| \\
    & + \sup_{f \in \mathcal{F}} \left|\hat{\mathrm{R}}_{\rm U}^{-} - \bar{\mathrm{R}}_{\rm U}^{-}\right|.
\end{align*}

It suffices to prove Lemma 1 if we can prove that with probability at least $1 - \frac{\delta}{k + 1}$, the following bounds hold separately:
\begingroup\makeatletter\def\f@size{9.5}\check@mathfonts
\def\maketag@@@#1{\hbox{\m@th\large\normalfont#1}}%
\begin{align}
    \sup_{f \in \mathcal{F}} \left|\hat{\mathrm{R}}_{{\rm{P}}_i}^{+} - \bar{\mathrm{R}}_{{\rm{P}}_i}^{+}\right| \leq {} & \frac{2L_l}{\tau}\mathfrak{R}_{n_{{\rm{P}}_i}, p(\boldsymbol{x} \mid y = i)}(\mathcal{F}) \nonumber\\
   & + \frac{(\tau + 1)C_l}{\tau}\sqrt{\frac{\log \frac{k + 1}{\delta}}{2n_{{\rm{P}}_i}}}, \label{pos}\\
    \sup_{f \in \mathcal{F}} \left|\hat{\mathrm{R}}_{\rm U}^{-} - \bar{\mathrm{R}}_{\rm U}^{-}\right| \leq {} &
    2 L_l \mathfrak{R}_{n_{\rm{U}}, p(\boldsymbol{x})}(\mathcal{F}) \nonumber\\
   & + C_l\sqrt{\frac{\log \frac{k + 1}{\delta}}{2n_{\rm U}}}. \label{unl}
\end{align}\endgroup

Next, we prove Inequation \ref{pos}. Inequation \ref{unl} is proven similarly.

Let $\phi_{\boldsymbol{x}}: \mathbb{R} \rightarrow \mathbb{R}$ be the function defined by $\phi_{\boldsymbol{x}}: z \mapsto \ell(z, i) + \ell(z, 0)\left(\mathds{1}_{\hat{\lambda}(\boldsymbol{x}) > \tau}\frac{1}{\hat{\lambda}(\boldsymbol{x})} - 1\right)$. For $\boldsymbol{x} \in \mathbb{R}^d$, $f \in \mathcal{F}$, since $\ell (\cdot) \in [0, C_l]$ and $\mathds{1}_{\hat{\lambda}(\boldsymbol{x}) > \tau}\frac{1}{\hat{\lambda}(\boldsymbol{x})} - 1 \in \left[-1, \frac{1}{\tau} - 1\right]$, we have $\phi_{\boldsymbol{x}}(f(x)) \in \left[-C_l, \frac{C_l}{\tau}\right]$. Following the proof of Theorem 3.3 in \cite{mohri2018foundations}, we can show that with probability at least $1 - \frac{\delta}{k + 1}$, it holds that
\begingroup\makeatletter\def\f@size{9}\check@mathfonts
\def\maketag@@@#1{\hbox{\m@th\large\normalfont#1}}%
\begin{align*}
    &\sup_{f \in \mathcal{F}} \left|\hat{\mathrm{R}}_{{\rm{P}}_i}^{+} - \bar{\mathrm{R}}_{{\rm{P}}_i}^{+}\right| \leq \\
    & 2 \mathbb{E}_{\mathcal{X}_{{\rm{P}}_i} \sim p(\boldsymbol{x} \mid y = i)^{n_{{\rm{P}}_i}}} \mathbb{E}_{\boldsymbol{\theta}}\left[\sup_{f\in \mathcal{F}}\frac{1}{n_{{\rm{P}}_i}}\sum_{j}^{n_{{\rm{P}}_i}}\theta_{j} \phi_{\boldsymbol{x}_{j}^{{\rm{P}}_i}}(f(\boldsymbol{x}_{j}^{{\rm{P}}_i}))\right] \\
    & + \frac{(\tau + 1)C_l}{\tau}\sqrt{\frac{\log \frac{k + 1}{\delta}}{2n_{{\rm{P}}_i}}},
\end{align*}\endgroup
where $\boldsymbol{\theta} = \{\theta_1, ..., \theta_{n_{{\rm{P}}_i}}\}$ and each $\theta_i$ is a Rademacher variable.

We notice that for all $\boldsymbol{x}$, $\phi_{\boldsymbol{x}}$ is a $(L_l/\tau)$-Lipschitz function on the interval $[-C_g, C_g]$. Following the proof of Lemma 26.9 in \cite{shalev2014understanding}, we can show that, when the set $\mathcal{X}_{{\rm{P}}_i}$ is fixed, we have
\begingroup\makeatletter\def\f@size{9}\check@mathfonts
\def\maketag@@@#1{\hbox{\m@th\large\normalfont#1}}%
\begin{align*}
    \mathbb{E}_{\mathcal{X}_{{\rm{P}}_i} \sim p(\boldsymbol{x} \mid y = i)^{n_{{\rm{P}}_i}}} \mathbb{E}_{\boldsymbol{\theta}}\left[\sup_{f\in \mathcal{F}}\frac{1}{n_{{\rm{P}}_i}}\sum_{j}^{n_{{\rm{P}}_i}}\theta_{j} \phi_{\boldsymbol{x}_{j}^{{\rm{P}}_i}}(f(\boldsymbol{x}_{j}^{{\rm{P}}_i}))\right] \leq \\
    \frac{L_l}{\tau}\mathbb{E}_{\mathcal{X}_{{\rm{P}}_i} \sim p(\boldsymbol{x} \mid y = i)^{n_{{\rm{P}}_i}}} \mathbb{E}_{\boldsymbol{\theta}}\left[\sup_{f\in \mathcal{F}}\frac{1}{n_{{\rm{P}}_i}}\sum_{j}^{n_{{\rm{P}}_i}}\theta_{j} f(\boldsymbol{x}_{j}^{{\rm{P}}_i})\right].
\end{align*}\endgroup
We can obtain Inequation \ref{pos} by substituting the Rademacher complexity. Lemma 1 is proven.\\

Then, we establish the error bound from $\bar{\mathrm{R}}(f)$ to $\mathrm{R}(f)$ by the following lemma.

\paragraph{Lemma 2.} \emph{Let $\hat{\lambda}(\cdot): \mathbb{R}^d \rightarrow (0, 1]$, $\tau \in (0, 1]$, $\zeta = p(\hat{\lambda}(\cdot) \leq \tau)$ and $\epsilon = \mathbb{E}_{\boldsymbol{x} \sim p(\boldsymbol{x})}\left[|\hat{\lambda}(\boldsymbol{x}) - \lambda(\boldsymbol{x})|^2\right]$. For all $f \in \mathcal{F}$, it holds that}
\begin{align*}
    \left|\bar{\mathrm{R}}(f) - \mathrm{R}(f)\right| \leq \frac{C_l}{\tau}\sqrt{(1 - \zeta)\epsilon}.
\end{align*}
\emph{Proof of Lemma 2.} We notice that the difference between $\bar{\mathrm{R}}(f)$ and $\mathrm{R}(f)$ is actually the difference between $\bar{\mathrm{R}}_{\rm U}^{-}$ and $\mathrm{R}_{\rm U}^{-}$, where 
\begingroup\makeatletter\def\f@size{9.5}\check@mathfonts
\def\maketag@@@#1{\hbox{\m@th\large\normalfont#1}}%
\begin{align*}
    \bar{\mathrm{R}}_{\rm U}^{-} = {} & \sum_{i=1}^{k} \pi_i \mathbb{E}_{\boldsymbol{x} \sim p(\boldsymbol{x} \mid y = i)} \left[\mathds{1}_{\hat{\lambda}(\boldsymbol{x}) > \tau} \ell(f(\boldsymbol{x}), 0)\frac{1}{\hat{\lambda}(\boldsymbol{x})}\right]\\
        & + \mathbb{E}_{\boldsymbol{x} \sim p(\boldsymbol{x})} \left[\mathds{1}_{\hat{\lambda}(\boldsymbol{x}) \leq \tau} \ell(f(\boldsymbol{x}), 0)\right],\\
    \mathrm{R}_{\rm U}^{-} = {} & \sum_{i=1}^{k} \pi_i \mathbb{E}_{\boldsymbol{x} \sim p(\boldsymbol{x} \mid y = i)} \left[\mathds{1}_{\lambda(\boldsymbol{x}) > \tau} \ell(f(\boldsymbol{x}), 0)\frac{1}{\lambda(\boldsymbol{x})}\right]\\
        & + \mathbb{E}_{\boldsymbol{x} \sim p(\boldsymbol{x})} \left[\mathds{1}_{\lambda(\boldsymbol{x}) \leq \tau} \ell(f(\boldsymbol{x}), 0)\right].
\end{align*}\endgroup
We can rewrite $\bar{\mathrm{R}}_{\rm U}^{-}$ and $\mathrm{R}_{\rm U}^{-}$ in the form of integral
\begingroup\makeatletter\def\f@size{9}\check@mathfonts
\def\maketag@@@#1{\hbox{\m@th\large\normalfont#1}}%
\begin{align*}
    \bar{\mathrm{R}}_{\rm U}^{-} = {} & \sum_{i=1}^{k} \pi_i \int \mathds{1}_{\hat{\lambda}(\boldsymbol{x}) > \tau} \ell(f(\boldsymbol{x}), 0)\frac{1}{\hat{\lambda}(\boldsymbol{x})}p(\boldsymbol{x} \mid y = i)d\boldsymbol{x} \\
    & + \int \mathds{1}_{\hat{\lambda}(\boldsymbol{x}) \leq \tau} \ell(f(\boldsymbol{x}), 0)p(\boldsymbol{x})d\boldsymbol{x},\\
    \mathrm{R}_{\rm U}^{-} = {} &  \sum_{i=1}^{k} \pi_i \int \mathds{1}_{\hat{\lambda}(\boldsymbol{x}) > \tau} \ell(f(\boldsymbol{x}), 0)\frac{1}{\lambda(\boldsymbol{x})}p(\boldsymbol{x} \mid y = i)d\boldsymbol{x}\\
    & + \int \mathds{1}_{\hat{\lambda}(\boldsymbol{x}) \leq \tau} \ell(f(\boldsymbol{x}), 0)p(\boldsymbol{x})d\boldsymbol{x},
\end{align*}\endgroup
where for $\mathrm{R}_{\rm U}^{-}$, we replace the subscript $\lambda(\boldsymbol{x})$ of the indicator function with $\hat{\lambda}(\boldsymbol{x})$, which does not change the value of $\mathrm{R}_{\rm U}^{-}$.

According to the sub-additivity of the supremum operator, we have
\begingroup\makeatletter\def\f@size{8.5}\check@mathfonts
\def\maketag@@@#1{\hbox{\m@th\large\normalfont#1}}%
\begin{align*}
    & \left|\bar{\mathrm{R}}(f) - \mathrm{R}(f) \right| = \left| \bar{\mathrm{R}}_{\rm U}^{-} - \mathrm{R}_{\rm U}^{-} \right| \leq \\
    & \sum_{i=1}^{k} \pi_i \int \mathds{1}_{\hat{\lambda}(\boldsymbol{x}) > \tau} \ell(f(\boldsymbol{x}), 0)\left| \frac{1}{\hat{\lambda}(\boldsymbol{x})} - \frac{1}{\lambda(\boldsymbol{x})} \right| p(\boldsymbol{x} \mid y = i)d\boldsymbol{x}\\
    &= \int \mathds{1}_{\hat{\lambda}(\boldsymbol{x}) > \tau} \ell(f(\boldsymbol{x}), 0)\left| \frac{1}{\hat{\lambda}(\boldsymbol{x})} - \frac{1}{\lambda(\boldsymbol{x})} \right| p(\boldsymbol{x}, y > 0)d\boldsymbol{x}\\
    &= \int \mathds{1}_{\hat{\lambda}(\boldsymbol{x}) > \tau} \ell(f(\boldsymbol{x}), 0) \frac{\left|\hat{\lambda}(\boldsymbol{x}) - \lambda(\boldsymbol{x})\right|}{\hat{\lambda}(\boldsymbol{x})\lambda(\boldsymbol{x})}  p(\boldsymbol{x}, y > 0)d\boldsymbol{x}\\
    & \leq \frac{C_l}{\tau} \int \mathds{1}_{\hat{\lambda}(\boldsymbol{x}) > \tau} \frac{\left|\hat{\lambda}(\boldsymbol{x}) - \lambda(\boldsymbol{x})\right|}{\lambda(\boldsymbol{x})}  p(\boldsymbol{x}, y > 0)d\boldsymbol{x}\\
    & = \frac{C_l}{\tau} \int \mathds{1}_{\hat{\lambda}(\boldsymbol{x}) > \tau} \frac{\left|\hat{\lambda}(\boldsymbol{x}) - \lambda(\boldsymbol{x})\right|}{p(y > 0 \mid \boldsymbol{x})}  p(y > 0 \mid \boldsymbol{x})p(\boldsymbol{x})d\boldsymbol{x}\\
    & = \frac{C_l}{\tau} \int \mathds{1}_{\hat{\lambda}(\boldsymbol{x}) > \tau} \left|\hat{\lambda}(\boldsymbol{x}) - \lambda(\boldsymbol{x})\right| p(\boldsymbol{x})d\boldsymbol{x}\\
    & \leq \frac{C_l}{\tau} \sqrt{\int \mathds{1}_{\hat{\lambda}(\boldsymbol{x}) > \tau}^2 p(\boldsymbol{x})d\boldsymbol{x}} \sqrt{\int \left|\hat{\lambda}(\boldsymbol{x}) - \lambda(\boldsymbol{x})\right|^2 p(\boldsymbol{x})d\boldsymbol{x}}\\
    & = \frac{C_l}{\tau}\sqrt{(1 - \zeta)\epsilon},
\end{align*}\endgroup
where the last inequality is obtained after applying the Cauchy-Schwarz inequality. Lemma 2 is proven.

Combining the above three error bounds, we know that with probability at least $1 - \delta$, the following holds: \useshortskip
\begingroup\makeatletter\def\f@size{8.5}\check@mathfonts
\def\maketag@@@#1{\hbox{\m@th\large\normalfont#1}}%
\begin{align*}
    &\sup_{f \in \mathcal{F}}\left|\hat{\mathrm{R}}_{\rm Conf-MPU}(f) - \mathrm{R}(f)\right| \leq 
    \sum_{i = 1}^{k} \pi_i \frac{(\tau + 1)C_l}{\tau}\\
    & + \sum_{i = 1}^{k}\pi_i \left[ \frac{2L_l}{\tau}\mathfrak{R}_{n_{{\rm{P}}_i}, p(\boldsymbol{x} \mid y = i)}(\mathcal{F}) 
     + \frac{(\tau + 1)C_l}{\tau}\sqrt{\frac{\log \frac{k + 1}{\delta}}{2n_{{\rm{P}}_i}}}\right] \\
     & + 2 L_l \mathfrak{R}_{n_{\rm{U}}, p(\boldsymbol{x})}(\mathcal{F}) + C_l\sqrt{\frac{\log \frac{k + 1}{\delta}}{2n_{\rm U}}}
     + \frac{C_l}{\tau}\sqrt{(1 - \zeta)\epsilon}.
\end{align*}\endgroup

Finally, with probability at least $1 - \delta$,
\begingroup\makeatletter\def\f@size{9}\check@mathfonts
\def\maketag@@@#1{\hbox{\m@th\large\normalfont#1}}%
\begin{align*}
    & \mathrm{R}(\hat{f}_{\rm Conf-MPU}) - \mathrm{R}(f^*)\\
    & = \mathrm{R}(\hat{f}_{\rm Conf-MPU}) - \hat{\mathrm{R}}_{\rm Conf-MPU}(\hat{f}_{\rm Conf-MPU}) \\
    &\quad + \hat{\mathrm{R}}_{\rm Conf-MPU}(\hat{f}_{\rm Conf-MPU}) - \hat{\mathrm{R}}_{\rm Conf-MPU}(f^*) \\
    &\quad + \hat{\mathrm{R}}_{\rm Conf-MPU}(f^*) - \mathrm{R}(f^*)\\
    & \leq \mathrm{R}(\hat{f}_{\rm Conf-MPU}) - \hat{\mathrm{R}}_{\rm Conf-MPU}(\hat{f}_{\rm Conf-MPU}) \\
    & \quad + \hat{\mathrm{R}}_{\rm Conf-MPU}(f^*) - \mathrm{R}(f^*)\\
    & \leq \left|\mathrm{R}(\hat{f}_{\rm Conf-MPU}) - \hat{\mathrm{R}}_{\rm Conf-MPU}(\hat{f}_{\rm Conf-MPU})\right| \\
    & \quad + \left| \hat{\mathrm{R}}_{\rm Conf-MPU}(f^*) - \mathrm{R}(f^*) \right|\\
    & \leq 2\sup_{f \in \mathcal{F}}\left|\hat{\mathrm{R}}_{\rm Conf-MPU}(f) - \mathrm{R}(f)\right|\\
    & \leq \sum_{i = 1}^{k} 2\pi_i \frac{(\tau + 1)C_l}{\tau}
     + \sum_{i = 1}^{k}2\pi_i \Bigg[ \frac{2L_l}{\tau}\mathfrak{R}_{n_{{\rm{P}}_i}, p(\boldsymbol{x} \mid y = i)}(\mathcal{F}) \\
     & \quad+ \frac{(\tau + 1)C_l}{\tau}\sqrt{\frac{\log \frac{k + 1}{\delta}}{2n_{{\rm{P}}_i}}}\Bigg] 
      + 4 L_l \mathfrak{R}_{n_{\rm{U}}, p(\boldsymbol{x})}(\mathcal{F}) \\
     & \quad + 2C_l\sqrt{\frac{\log \frac{k + 1}{\delta}}{2n_{\rm U}}}
     + \frac{2C_l}{\tau}\sqrt{(1 - \zeta)\epsilon}.
\end{align*}\endgroup
Theorem \ref{theorem 1} is proven.



\section{Evaluation of Prior Estimation}
As mentioned in Section \ref{sec:prior}, we apply the TIcE algorithm to estimate the prior for each class without using ground-truth annotations. In Table \ref{tab:prior}, we compare the priors estimated on distantly annotated BC5CDR (Big Dict) and CoNLL2003 (Dict) with the true priors for each class of the two datasets. We also conduct the prior estimation on BC5CDR (Small Dict) and CoNLL2003 (KB), and we find that the priors estimated with different dictionaries are not significantly different.
Table \ref{tab:prior} shows that the estimated priors by TIcE algorithm are close to the true priors. We further trained Conf-MPU models using the true priors and do not observe significant differences in the performances. This experiment indicates that Conf-MPU is not sensitive to the prior estimations and TIcE algorithm can be applied for prior estimation without ground-truth labels. 

\begin{table}[H]
\centering

\resizebox{0.48\textwidth}{!}{%
\begin{tabular}{clcc}
\Xhline{2 \arrayrulewidth}
Dataset                    & Type     & Estimated Prior & True Prior \\
\Xhline{2\arrayrulewidth}
\multirow{2}{*}{\begin{tabular}[c]{@{}c@{}}\textbf{BC5CDR}\\\textbf{(Big Dict)}\end{tabular}}    & Chemical & 0.0503          & 0.0601     \\
                          & Disease  & 0.0504          & 0.0601     \\
\Xhline{1.5\arrayrulewidth}
\multirow{4}{*}{\begin{tabular}[c]{@{}c@{}}\textbf{CoNLL2003}\\\textbf{(Dict)}\end{tabular}} & PER      & 0.1052          & 0.0547     \\
                          & LOC      & 0.0331          & 0.0407     \\
                          & ORG      & 0.0630          & 0.0492     \\
                          & MISC     & 0.0371          & 0.0226    \\
\Xhline{2\arrayrulewidth}
\end{tabular}%
}
\caption{The results of prior estimation.}
\label{tab:prior}
\end{table}

\section{Evaluation of Confidence Score Estimation}
For Conf-MPU, another factor for its performance is the confidence score estimation for each token being an entity token. To evaluate the quality of the confidence scores, we first convert the results as labels where if $\hat{\lambda}(\boldsymbol{x})>0.5$ then label $\boldsymbol{x}$ as an entity token, otherwise label as a non-entity token. 
We present the results in terms of token-level F1 score, precision, and recall in Table \ref{tab:first-step}, where Supervised PN using human-annotated ground-truth labels provides upper-bound references for this estimation.
We can see that the classifier with the binary PU risk estimation achieves good recall on all of the distantly labeled datasets. High recall indicates that the classifier can recognize most of entity tokens, which can be taken advantage of in the Conf-MPU risk estimation to avoid overfitting to false negative samples in unlabeled data. We also evaluated the proposed Conf-MPU models with the confidence scores given by the supervised PN classifier on the four distantly labeled datasets, where the performances increased by 2 $\sim$ 5 percentage in terms of F1 score, indicating that the proposed Conf-MPU framework is robust to the confidence score estimation of lesser quality. We leave for future work the optimization of the confidence score estimation.

\begin{table}[]
\centering
\resizebox{0.48\textwidth}{!}{%
\begin{tabular}{lcc}
\Xhline{2\arrayrulewidth}
Method      & \textbf{BC5CDR (Big Dict)}     & \textbf{BC5CDR (Small Dict)}  \\
\Xhline{1.5\arrayrulewidth}
Supervised PN    & \multicolumn{2}{c}{85.00 (77.68/93.83)}      \\
Binary PU   & 72.38 (61.89/87.17)   & 79.14 (81.85/76.60)  \\
\hline \hline
Method      & \textbf{CoNLL2003 (KB)}        & \textbf{CoNLL2003 (Dict)}     \\
\Xhline{1.5\arrayrulewidth}
Supervised PN    & \multicolumn{2}{c}{96.09 (92.77/99.65)}      \\
Binary PU   & 88.88 (81.04/98.39)   & 88.08 (79.09/99.38)  \\
\Xhline{2\arrayrulewidth}
\end{tabular}%
}
\caption{The results of confidence score estimation on test sets: F1 (Precision/Recall) (in \%).}
\vspace{3mm}
\label{tab:first-step}
\end{table}

\begin{figure}[]
    \centering
    \includegraphics[width=0.48\textwidth]{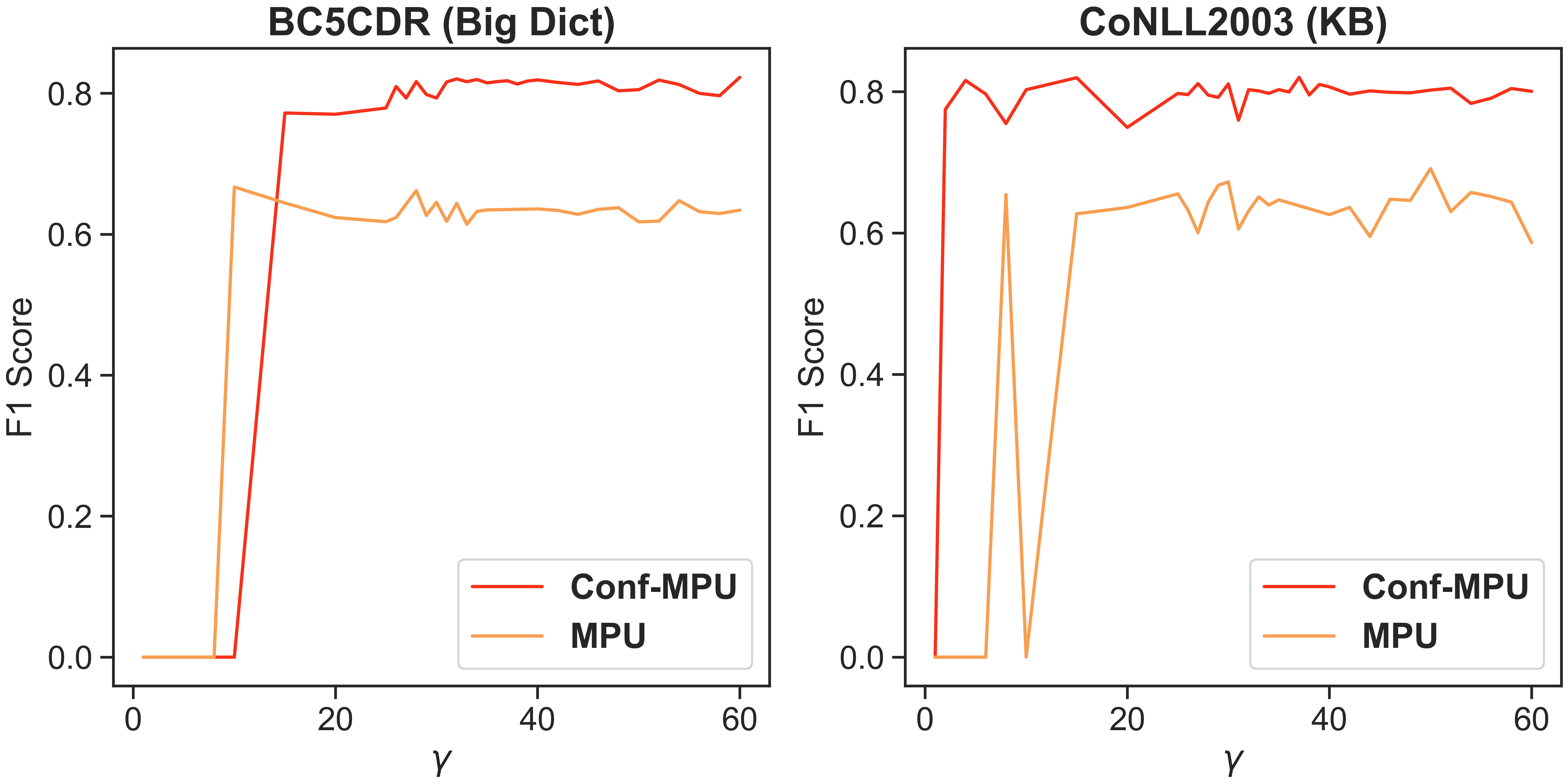}
    \caption{Empirical study on the class weight.}
    \label{class weight}
\end{figure}

\section{Study on Class Imbalance Problem} NER tasks often suffer from the problem of class imbalance, where most tokens are not entity tokens. In our experiments, we introduce a hyper-parameter $\gamma$ as a class weight in risk estimations to balance the risks on positive and unlabeled data.

We empirically investigate the effect of this parameter by evaluating Conf-MPU and MPU with different values of $\gamma$ on BC5CDR (Big Dict) and CoNLL2003 (KB), and show the span-level F1 scores on test sets in Figure \ref{class weight}. We can see that if the class imbalance issue is ignored (\textit{i.e.}, $\gamma = 1$), the two methods achieve very low F1 scores on the two distantly labeled datasets. When $\gamma$ increases to a certain value (\textit{e.g.}, $\gamma = 15$), both methods can achieve good performances. As $\gamma$ increases, the F1 scores fluctuate a little within a certain range but stay high. It indicates that the two methods are not sensitive to the value of the class weight $\gamma$ if it is properly large. Similar results can also be observed on the other distantly labeled datasets. Therefore, for a fair comparison, we uniformly set $\gamma$ to 28 and 15 for distantly labeled BC5CDR datasets and CoNLL2003 datasets, respectively, in our experiments.

\endgroup

\end{document}